\documentclass[conference,letterpaper,10pt]{IEEEtran}
\IEEEoverridecommandlockouts
\usepackage{amsmath}
\usepackage{mathtools}
\usepackage{color}
\usepackage{array}
\usepackage{stfloats}
\usepackage{amsthm}
\usepackage{amssymb}
\usepackage{float}
\usepackage{nccmath}
\usepackage{graphicx}
\usepackage[linesnumbered,ruled]{algorithm2e}
\usepackage{setspace}
\usepackage{booktabs}
\usepackage{subcaption}
\usepackage{mwe}
\usepackage{cite}
\usepackage{amsmath,amssymb,amsfonts}
\usepackage{graphicx}
\usepackage{textcomp}
\usepackage{xcolor}

\usepackage{algorithmicx}
\usepackage{url}
\usepackage{xspace}
\usepackage{graphicx}  
\usepackage{subcaption}

\def\BibTeX{{\rm B\kern-.05em{\sc i\kern-.025em b}\kern-.08em
    T\kern-.1667em\lower.7ex\hbox{E}\kern-.125emX}}
\begin{document}

\title{Visualization of Deep Reinforcement Autonomous Aerial Mobility Learning Simulations}

\author{
\IEEEauthorblockN{Gusang Lee}
\IEEEauthorblockA{\textit{Korea University}\\
rntkd0917@korea.ac.kr}
\and
\IEEEauthorblockN{Won Joon Yun}
\IEEEauthorblockA{\textit{Korea University}\\
ywjoon95@korea.ac.kr}
\and
\IEEEauthorblockN{Soyi Jung}
\IEEEauthorblockA{\textit{Korea University}\\
jungsoyi@korea.ac.kr}
\and
\IEEEauthorblockN{Joongheon Kim}
\IEEEauthorblockA{\textit{Korea University}\\
joongheon@korea.ac.kr}
\and
\IEEEauthorblockN{Jae-Hyun Kim}
\IEEEauthorblockA{\textit{Ajou University}\\
jkim@ajou.ac.kr}
}
\maketitle 

\begin{abstract}
This demo abstract presents the visualization of deep reinforcement learning (DRL)-based autonomous aerial mobility simulations. In order to implement the software, Unity-RL is used and additional buildings are introduced for urban environment. On top of the implementation, DRL algorithms are used and we confirm it works well in terms of trajectory and 3D visualization.
\end{abstract}

\section{Introduction}\label{sec:1}
As a major part of beyond 5G (B5G) or 6G network scenarios, autonomous urban aerial mobility (UAM) systems are widely and actively discussed by industry and academia~\cite{nm20saad}. Based on these huge interests, many research contributions are available nowadays in terms of UAM trajectory optimization~\cite{tvt19yin}, energy-efficient operations~\cite{tvt19shin}, and so forth.
The trajectory optimization and energy-efficient operations fundamentally control the mobility of autonomous UAM systems. 
Therefore, conducting visual simulations with the proposed learning-based trajectory optimization and energy-efficient operation algorithms is essentially required in order to intuitively understand the behaviors of the UAM algorithms. 
In addition, the most of trajectory optimization algorithms are designed via deep reinforcement learning (DRL) because DRL algorithms are fundamentally for sequential stochastic decision making in order to maximize cumulative expected rewards. Therefore, the UAM simulations should efficiently identify the DRL-based autonomous UAM flying trajectories and operations, and thus, it is obvious that visual representation of the UAM simulations is helpful for intuitive understanding the algorithms. 

In this paper, we implement our own 3D visualization software platform which is for simulating DRL-based autonomous trajectory control using Unity. In order to conduct more precise simulations, we added buildings for urban environment because we assume urban aerial mobility for providing smart city services such as surveillance and flexible mobile access.  


\section{Unity Implementation and 3D Visualization}\label{sec:2}

\subsection{Unity Implementation}
\begin{figure}[t!]
    \centering
    \includegraphics[width=0.81\columnwidth]{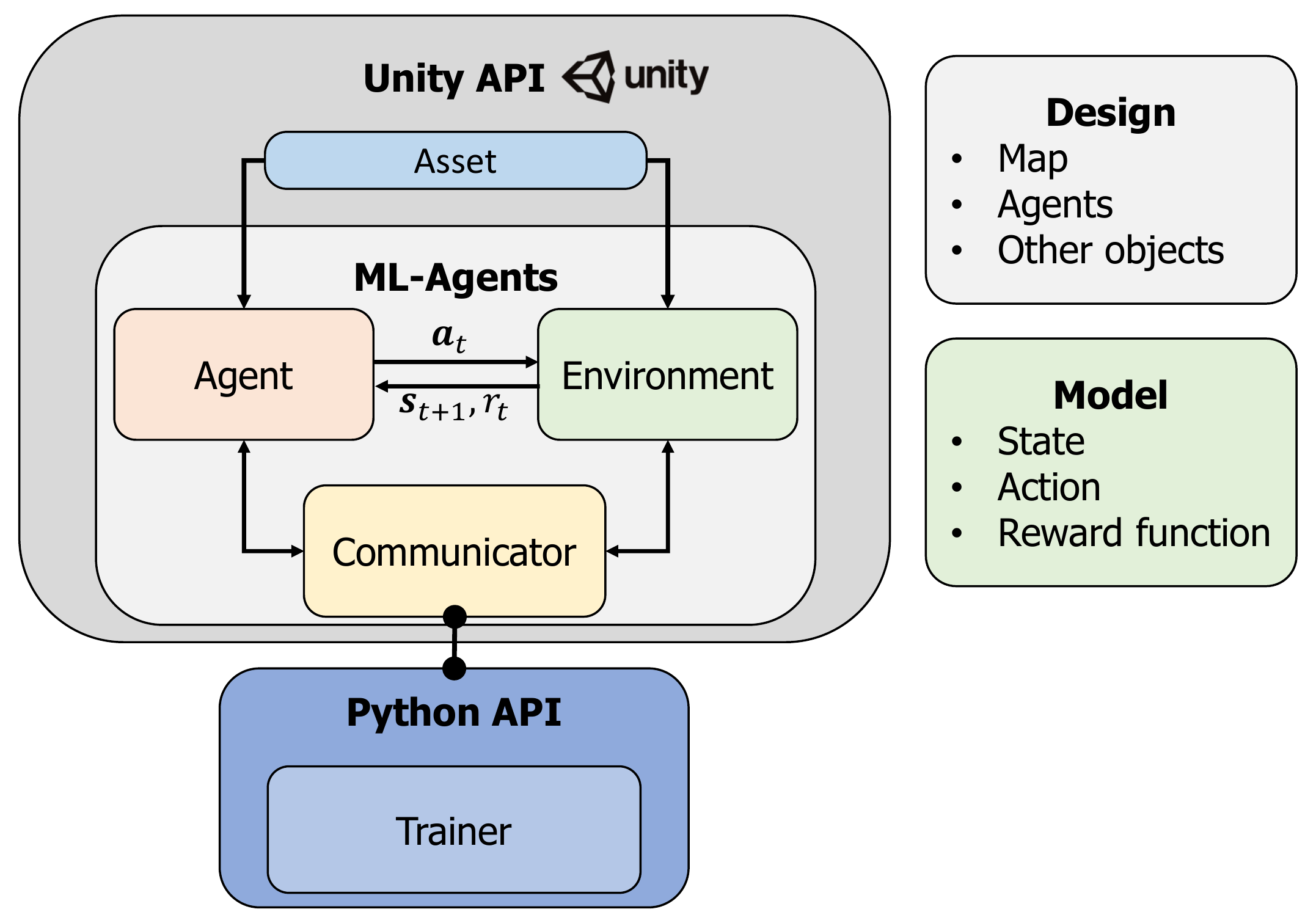}
    \caption{Software architecture for learning environment.}
    \label{fig:1}
    \vspace{-2mm}
\end{figure}

Fig.~\ref{fig:1} illustrates the system architecture for conducting DRL in Unity environment. With Unity API, it is possible to model the environment, dynamic models and features, DRL elements (\textit{i.e.}, states, actions, transitions, and rewards), and these are named to \textit{Asset} in Unity. 
With \texttt{mlagents} (\textit{i.e.}, Unity library for DRL implementation), training DRL agents and visualizing the training results can be realized because 1) \textit{Communicator} exists which realizes the interaction with Python API and 2) \textit{Asset} can be loaded. For the simulations of UAM systems, Unity \textit{Asset} which is named to \textit{Drone Flight} is used. 

\begin{figure*}[t!]
    \centering
    \includegraphics[width=0.8\linewidth]{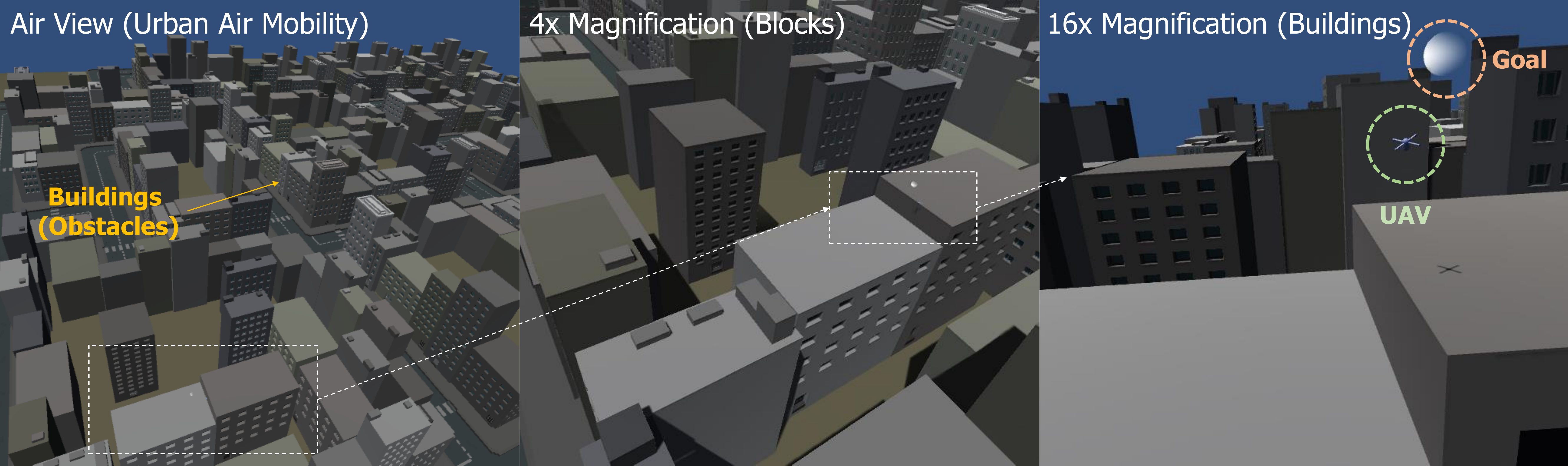}
    \caption{Unity implementation in UAM environment.}
    \label{fig:2}
    \vspace{-2mm}
\end{figure*}

Our considering aerial mobility system is UAM, thus the simulations should be performed in urban areas those are with numerous building and skyscrapers. Therefore, we implements the buildings and skyscrapers in \textit{Drone Flight}. Furthermore, the corresponding environment information which can be observed by agents is organized by current position, goal position, current velocity, current angular velocity, altitude vector, and building/skyscraper position vectors. The actions are for controlling the UAM motor for desired directions, thus 3D Cartesian coordinate is used, \textit{i.e.}, $(x,y,z)$. Lastly, the rewards can be positive when 1) the agent arrives at the destination whereas the rewards are negative when 1) the agent becomes far from the 
goal and 2) the agent becomes closer to buildings, skyscrapers, and obstacles. 
Fig.~\ref{fig:2} shows the Unity implementation results in this UAM environment. Note that several buildings are added for urban scenario construction. 

\subsection{Visualization}
Based on our Unity implementation on top of \textit{Drone Flight}, we conduct DRL-based agent trajectory training and performance evaluation. In addition, the results are visualized. 
For the DRL training of the agent, proximal policy optimization (PPO) is used~\cite{ppo}. The policy of agent is trained by the deep neural networks with 2 dense layers where each layer is with 128 units. In addition, $\epsilon$-greedy is used for DRL training exploration where $\epsilon=0.2$. 
Furthermore, multi-agent parallel processing is utilized that is supported by \texttt{mlagents}, thus parallel accelerated training computation is realized. For the DRL training, learning iteration is set to $3,000$\,K and detailed hardware/software specification is summarized in Table~\ref{tab:param1}.

\begin{table}[t]
\caption{Hardware and software specification}
\footnotesize
\label{tab:param1}
\begin{center}
	\centering
	\begin{tabular}{l|l}
    \toprule[1.0pt]
    \centering
      System  & Specification \\
    \midrule[1.0pt]
    CPU & $\bullet$ Intel(R) Core(TM) i7-9700k, 3.60GHz@2 \\
        & \hspace{1mm} RAM: 64\,GB \\ 
    \midrule
    GPU & $\bullet$ NVIDIA GeForce GTX 1660 super \\ 
        & \hspace{1mm} The number of cores: $1,408$ \\
        & \hspace{1mm} Frame buffer: 6GB GDDR6 \\
        & \hspace{1mm} Memory speed: 14 Gbps \\
    \midrule
    Software  & $\bullet$ Unity: 20.1.17f1 \\
    Ver.          & $\bullet$ \texttt{mlagents}: 0.7 \\
              & $\bullet$ tensorflow: 1.12 \\
    \bottomrule[1.0pt]
	\end{tabular}
\end{center}
\end{table}


\begin{figure}[t!]
    \centering
    \includegraphics[width=0.9\columnwidth]{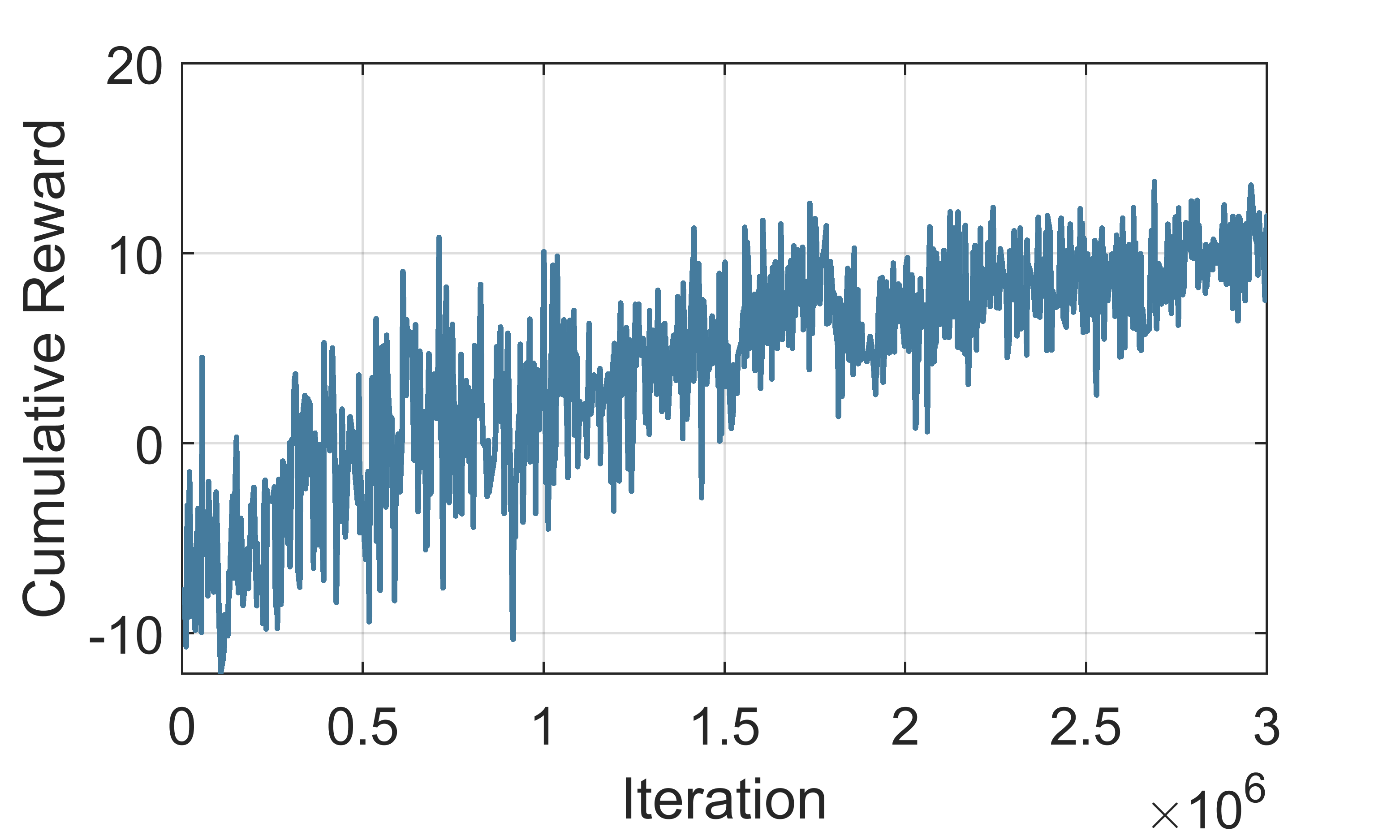}
    \caption{Rewards of autonomous aerial mobility learning.}
    \label{fig:3}
    \vspace{-2mm}
\end{figure}

\begin{figure}[t!]
    \centering
    \includegraphics[width=0.8\columnwidth]{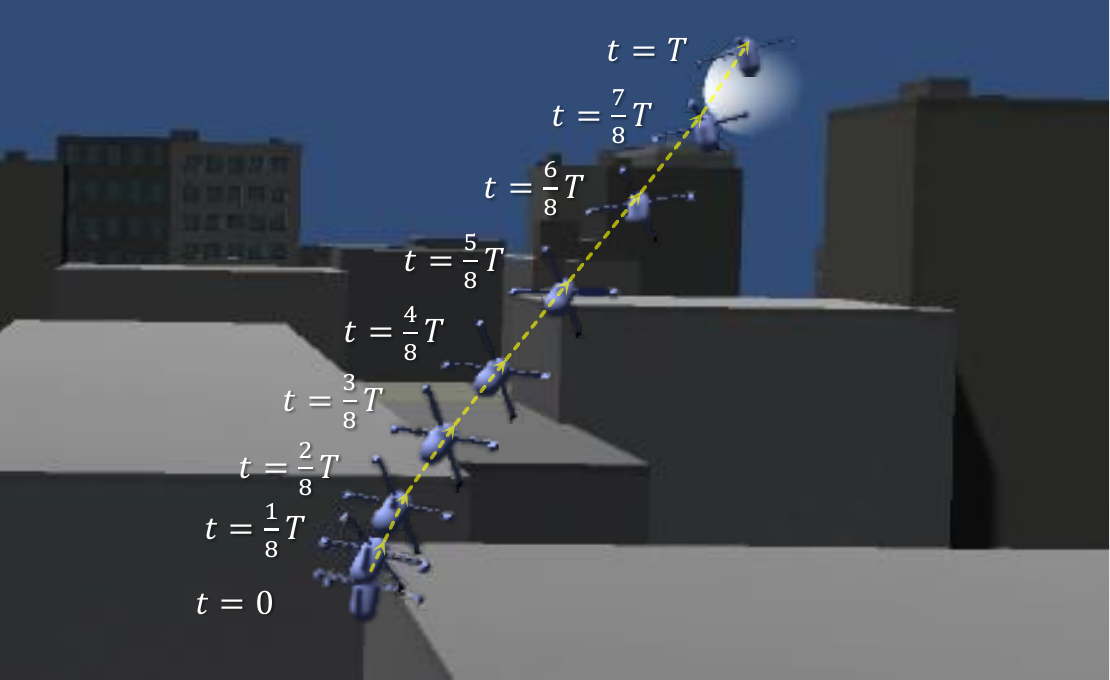}
    \caption{Visualization of autonomous aerial mobility learning.}
    \label{fig:4}
    \vspace{-2mm}
\end{figure}

Fig.~\ref{fig:3} and Fig.~\ref{fig:4} show the simulation results. In Fig.~\ref{fig:3}, the reward convergence is plotted, and then we can confirm that the reward eventually converges. 
Fig.~\ref{fig:4} shows the time-series visual simulations of DRL-trained UAM agent's behaviors.
Then, it can be observed that 1) the UAM agent moves toward its own goal when $t \in [0, T]$ and 2) the UAM tries to avoid buildings, skyscrapers, and obstacles when $t \in [0, \frac{2}{8}T]$, as designed in rewards. Note that the behaviors to move to the goal can be observed during all time steps. 
%
%

    

Finally, we observe that the DRL reward converges and the corresponding DRL-based agent controls its own trajectory based on the positive and negative reward settings. Therefore, we can confirm that our DRL-based agent works as desired and the results are simulated and visualized via our own Unity-based software visual simulation platforms.

Note that the video demonstration for our own simulation and 3D visualization results are in~\cite{youtube}.

\section{Conclusions and Future Work}\label{sec:3}
This demo abstract presents the implementation and visualization of DRL-based autonomous UAM simulations. Furthermore, various buildings can be placed for smart city urban environment simulations. As future work, various urban scenarios can be also considerable. 

\section*{Acknowledgment}
This research is supported by National Research Foundation of Korea (2019R1A2C4070663 and 2019M3E4A1080391).
S. Jung, J. Kim, and J.-H. Kim are corresponding authors.


\begin{thebibliography}{1}
	
	\bibitem{nm20saad}
	W. Saad, M. Bennis, and M. Chen, ``A vision of 6G wireless systems: Applications, trends, technologies, and open research problems," \textit{IEEE Network}, vol. 34, no. 3, pp. 134--142, May/June 2020.
	
	\bibitem{tvt19yin} 
	S. Yin, S. Zhao, Y. Zhao, and F. R. Yu, ``Intelligent trajectory design in UAV-aided communications with reinforcement learning," \textit{IEEE Trans. Veh. Technol.}, vol. 68, no. 8, pp. 8227--8231, August 2019.
	
	\bibitem{tvt19shin} 
	M. Shin, J. Kim, and M. Levorato, ``Auction-based charging scheduling with deep learning framework for multi-drone networks," \textit{IEEE Trans. Veh. Technol.}, vol. 68, no. 5, pp. 4235--4248, May 2019.
	
	\bibitem{ppo} 
	J. Schulman, F. Wolski, P. Dhariwal, A. Radford, and O. Klimov, ``Proximal policy optimization algorithms," \textit{CoRR},  vol. abs/1707.06347, 2017. [Online]. Available: \url{http://arxiv.org/abs/1707.06347}
	
	\bibitem{youtube} G. Lee, W. J. Yun, S. Jung, J. Kim, and J.-H. Kim, ``DRL-based autonomous UAM simulations and visualization," 2021. [Online]. Available: \url{https://sites.google.com/view/drl-uav-visualization}
	
\end{thebibliography}

\end{document}